\documentclass[preprint,12pt]{elsarticle}
\usepackage[numbers, sort&compress, super]{natbib}
\usepackage{amssymb}
\usepackage{xcolor}%

\newcommand\zfig[1]{{\color{violet}#1}}
\usepackage{times}
\usepackage{amsmath,amssymb,amsfonts}%
\usepackage{amsthm}%
\usepackage{mathrsfs}%
\usepackage{hyperref} 
\usepackage{caption}
\usepackage{subcaption}
\usepackage{geometry}

\begin{document}

\begin{frontmatter}

\title{Predicting Fatigue Crack Growth via Path Slicing and Re-Weighting}

\author[1]{Yingjie Zhao}
\author[1]{Yong Liu}

\affiliation[1]{organization={Applied Mechanics Laboratory, Department of Engineering Mechanics, Tsinghua University},
            city={Beijing},
            postcode={100084}, 
            country={China}}

\author[1,2]{Zhiping Xu} 
\affiliation[2]{Lead contact. Correspondence: xuzp@tsinghua.edu.cn (Z.X.)}

\begin{abstract}
Predicting potential risks associated with the fatigue of key structural components is crucial in engineering design.
However, fatigue often involves entangled complexities of material microstructures and service conditions, making diagnosis and prognosis of fatigue damage challenging.
We report a statistical learning framework to predict the growth of fatigue cracks and the life-to-failure of the components under loading conditions with uncertainties.
Digital libraries of fatigue crack patterns and the remaining life are constructed by high-fidelity physical simulations.
Dimensionality reduction and neural network architectures are then used to learn the history dependence and nonlinearity of fatigue crack growth.
Path-slicing and re-weighting techniques are introduced to handle the statistical noises and rare events.
The predicted fatigue crack patterns are self-updated and self-corrected by the evolving crack patterns.
The end-to-end approach is validated by representative examples with fatigue cracks in plates, which showcase the digital-twin scenario in real-time structural health monitoring and fatigue life prediction for maintenance management decision-making.
\end{abstract}

%

\end{frontmatter}


\clearpage
\newpage

\section*{Introduction}

\noindent Fatigue life prediction (FLP) is of critical importance for structural integrity design in, for example, aerospace and nuclear engineering~\cite{schijve2009fatigue}.
After fatigue initiation with accumulated damage, fatigue cracks grow and can be monitored after the size reaches a detection threshold.
In practice, periodic inspection is commonly arranged to identify flaws in the structural components.
The information is fed into fracture mechanics analysis (FMA) where the remaining life can be calculated from empirical rules of fatigue crack growth (FCG).
Predictive maintenance schemes could further reduce the life-cycle cost and increase system safety, which have been actively explored in recent studies~\cite{kapteyn2021probabilistic}.

However, the physics governing fatigue is entangled with the microstructural evolution of materials and the profiles of loading conditions~\cite{zhang2022JMPS,stinville2022origins}.
The microscopic processes of fatigue may involve plasticity, fracture, and phase transitions, which are defined by chemical compositions, and atomic-level and microscopic structures~\cite{xu2023environmentally}.
As a result, material fatigue, like fluid turbulence, becomes a complex system process of the microscopic components with their interaction spanning across multiple space and time scales~\cite{lambiotte2019NatutePhys}.
The nonlinearity and heterogeneity embedded in the mathematical or data-driven models make predicting behaviors of these systems challenging~\cite{runge2019SciAdv}.
Fingerprint features of materials and structural components as well as the history or path dependence of FCG make their responses susceptible to statistical noises and rare events resulting from intrinsic or extrinsic sources~\cite{kishore2011PRL,qi2020PNAS}.
Uncertainties thus exist in the microstructure-sensitive constitutive relations of materials and the loading conditions in experimental tests or under specific service conditions, respectively, which can alter the processes of material damage and FCG~\cite{sankararaman2011EFM}.

Material responses during fatigue can be characterized by statistical data of the fatigue life or FCG rates obtained from experimental tests~\cite{zhang2023fatigue_am,zhang2023fatigue_cma}.
With this knowledge, recent developments in data science and machine learning techniques allow engineers to take a data-driven approach to structural health monitoring (SHM) and FLP~\cite{farrar2012SHM}.
However, a practical solution that tackles the history dependence, statistical noises, and rare events has not been established yet~\cite{do2019EFM,yang2021IJF}.
With digital libraries constructed from high-fidelity physical simulations, neural networks with specially designed architectures can extract the characteristic features and make predictions~\cite{mozaffar2019PNAS, buehler2022BEA}.
The history dependence can be handled by the long short-term memory (LSTM) network by introducing gating and memory functions~\cite{hochreiter1997LSTM}.
Surrogate models trained by machine learning techniques can reduce the computational costs significantly from physical modeling.
Prediction in a real-time or digital-twin paradigm can thus be achieved, where the interaction with physical sensor networks (PSNs) can be included to update the model parameters, quantify the uncertainties, and evaluate the models based on the Bayesian theory~\cite{box2011bayesian,niederer2021NatComputSci}.
 
In this work, we develop a statistical learning framework to predict the growth of fatigue cracks and the life-to-failure of structural components.
Digital libraries are constructed by finite element analysis (FEA).
Variational autoencoder (VAE) is used to learn the latent representation of the fatigue cracks, followed by LSTM and feedforward neural network (FNN) for the history dependence and the life prediction of FCG, respectively.
Path slicing and re-weighting techniques are introduced to address uncertainties in the service conditions.
Fully resolved fatigue crack paths and accurate prediction of the remaining life are demonstrated by examples.

\section*{Results}

\subsection*{The statistical learning framework.}
\noindent In engineering, FLP can be achieved through techniques that can be expressed into $4$ levels \zfig{(Figure 1)}.
Empirical models for the relations between the FCG rate (${\rm d}a/{\rm d}N$) and the amplitude of stress intensity factors (SIFs, $\Delta K$) can be fed into analytical solutions of SIF under specific loading conditions and sample geometries~\cite{al1998SIFs_handbook}.
FEA calculates the evolution of SIFs as fatigue cracks grow with high accuracy.
The surrogate models constructed using, for example, statistical learning methods can reduce the computational costs.
Digital twins with accurate and fast algorithms of FLP and interaction with the physical systems can be implemented for real-time monitoring and prognosis.
Our FLP framework includes modules of data generation, model training, and applications \zfig{(Figure 2, see Experimental Procedures for details)}.
The extended finite element method (XFEM) is utilized for modeling and constructing digital libraries.
Statistical noises and rare events are considered through the loading amplitudes with Gaussian distributions in FCG modeling using XFEM.
The produced datasets contain fatigue cracks and their corresponding residual life.
We combine VAE, LSTM, and FNN in model training, which is used for SHM and FLP for the structural components in the application module.
A path-slicing technique is used based on the datasets produced by FEA for the history dependence and statistical noises, and a re-weighing technique is introduced in the training process to signify the impact of rare events on the model parameters \zfig{(Figure 3A)}.

\subsection*{Dimensionality reduction of the crack patterns.}
\noindent Crack patterns or fractured surfaces modeled by XFEM are stored as voxel data.
These data characterize the competition between the material resistance to fracture and the driving force of FCG. 
For simple structures, the mapping between the far-field loading conditions and the crack-tip driving forces can be captured by FMA.
As a result, dimensionality reduction of the crack patterns is performed to extract crack surfaces to improve the learning efficiency in downstream tasks.
We use VAE to learn the latent representations of fatigue cracks in reduced dimensions.
The density distributions of fatigue crack patterns spanned in the space of the two primary features are summarized in \zfig{Figure 3B}.
The latent representations of fatigue cracks tend to follow the Gaussian distribution owing to the competition between the reconstruction loss and the Kullback-Leibler (KL) divergence in VAE \zfig{(see Experimental Procedures for details)}~\cite{kingma2013VAE}.

\subsection*{Path slicing and re-weighting for history dependence, statistical noises, and rare events.}
\noindent We introduce a path-slicing technique to address the complexity of the loading profiles and the effects of statistical noises on FCG \zfig{(Figure 3A)}.
The domain of interest is discretized to $N_{\rm s}$ regions.
As FCG proceeds across their boundaries and reaches the next region, the tension and shear loads are re-sampled from the Gaussian distribution \zfig{(Figure 3A)}.
Structural components in service may also experience rare events such as catastrophic failure of the components and unexpected external impact.
Rare events are identified from the fatigue crack patterns in the pre-constructed digital libraries by unsupervised clustering in the latent space (shown as red data points in the space spanned by two major features in \zfig{Figure 3C})~\cite{schubert2017dbscan}.
Their weights are boosted in downstream training to improve the performance of prediction, where an enrichment factor $\lambda$ is defined for the updated loss function
\begin{equation}\label{eq1}
{\rm Loss} = {\rm MSE}(\mathbf{z},\mathbf{\widehat{z}})+\lambda{\rm MSE}(\mathbf{z}_{\rm rare},\mathbf{\widehat{z}}_{\rm rare}),
\end{equation}
where $\mathbf{z}$ and $\mathbf{\widehat{z}}$ are the ground truth and predictions of the latent vectors, respectively.
The subscript indicates the rare records of fatigue cracks.

\subsection*{Prediction and correction in digital twins.}
\noindent To demonstrate the capability of our framework, we consider a flat plate subjected to loads with statistical noises and rare events as a representative example (\zfig{Figure 4}).
\zfig{Figures 4A-4C} show the representative fatigue crack paths from FLP.
The results show that without path slicing or re-weighting, the deflection of fatigue cracks cannot be correctly predicted.
FCG along the direction of the embryo crack is predicted without necessary changes induced by the uncertainties in the loading conditions \zfig{(Figure 4A)}.
In contrast, path slicing solves the problem by constantly updating the model prediction based on observation in experiments or physical modeling (in our work) at time $t_{\rm obs}$, correctly predicting crack deflection \zfig{(Figure 4B)}.
With re-weighting further implemented, the process of FCG and the remaining life are accurately predicted \zfig{(Figure 4C)}.

For quantitative assessment of the framework, the root mean square error (RMSE) and structural similarity (SSIM) are evaluated between the predicted crack path and the ground truth as a function of the observation time, $t_{\rm obs}$ \zfig{(Figure 4D, see Experimental Procedures for details)}.
RMSE offers a local measure for the deviation of fatigue crack paths, showing that combining path slicing and re-weighting effectively minimizes the errors by addressing the history dependence and noisy features in the loading conditions as well as the presence of rare events.
SSIM measures the global similarity from the mean and variance of the voxel values, showing that the predictions of fatigue crack paths are improved as $t_{\rm obs}$ increases \zfig{Figures 5C and 5D} show the error in life prediction as well as the accuracy of representative paths and all the path samples in the test set, respectively, which comprises $20\%$ of data in the digital libraries.
As $t_{\rm obs}$ increases, the performance of models equipped with path slicing and re-weighting is continuously improved.

In practice, the loading conditions can change continuously in service, the complexity of which is quantified using symbolic aggregate approximation (SAX) \zfig{(Figures 6A-6C, see Experimental Procedures for details)}~\cite{lin2003SAX}.
We find that path slicing and re-weighting can significantly reduce the data complexity actually needed in the statistical learning framework.
Specifically for the plate example, with only $200$ loading-profile samples in the training set, path slicing with $N_{\rm s} = 5$ can accurately predict FCG in the test set with $200$ samples, even for the loading conditions with statistical noises and rare events \zfig{(Figures 6B and 6C)}.
It should be noted that the performance of prediction depends on the size of training sets and the choice of $N_{\rm s}$.
Their values are determined for structural components for specific geometries, sizes, and loading conditions, and can be guided by the analysis of the loading-profile complexity.


\section*{Discussion and Conclusion}
\noindent To predict FCG and the life-to-failure in real time, our digital framework needs only the information of fracture crack morphologies, which can be integrated with PSNs and FMA to assess the model performance~\cite{qi2021SHM}.
In addition to crack patterns captured by high-speed cameras, PSN data such as local strain measured by strain gauges, elastic waves detected by acoustic emission sensors, and electrical impedance measured by piezoelectric sensors can be used to identify or infer the features of fatigue cracks and loading conditions~\cite{yoon2022SR}.
These data can be fed into FMA to calculate the stress fields at the crack tips based on analytical models or FEA~\cite{qi2021SHM}.
FCG is then predicted by using, for example, the Paris-Erdogan equation, and can be used to evaluate the accuracy of our statistical learning framework.
In practice, implementing digital twins for FLP still faces challenges in addition to real-time monitoring of fatigue crack patterns.
Firstly, the use of sensors on critical structural components such as turbine blades in the engines may not be feasible in a harsh environment.
Even for structural components where sensors can be installed, the driving force of FCG at the crack tip needs to be calculated from limited data from the PSNs.
Data-driven approaches were recently proposed to address this issue.
Strain fields are predicted from data collected from digital image correlation (DIC) and a limited number of strain gauge sensors using deep neural networks and experimental libraries~\cite{yoon2022SR}.
Secondly, FMA using FEA, although offering higher accuracy compared to analytical models, demands a considerable amount of time for numerical calculations, especially for structural components with complex geometries, crack-tip morphologies, and load conditions \zfig{(Figure 6D)}.
The applications of our model in FLP can be extended straightforwardly to engineering components such as the turbine blades with a 3D geometry, where the surface cracks are represented by using the point-cloud representation~\cite{anvekar2022CVPR,zhang2021AAAI}.
However, the digital models need to be updated by the observed physical states typically at computing time scale within milliseconds.
Pre-calculated digital libraries and surrogate models mitigate this burden, although including extremely rare events that deviate far from the Gaussian distribution (e.g., bird strikes on aircraft) remains difficult.
Our end-to-end, self-updated model thus offers FLP from the evolution of fatigue crack patterns and can interact with PSNs for performance.
The demonstrated accuracy and efficiency by considering the history dependence, statistical noises, and rare events can be integrated into digital twins of aerospace and nuclear power applications.


\section*{Experimental procedures}

\subsection*{Resource availability}

\subsubsection*{Lead contact}
\noindent Further information and requests for resources should be directed to and will be fulfilled by the lead contact, Zhiping Xu (xuzp@tsinghua.edu.cn).

\subsubsection*{Materials availability}
\noindent This study did not generate new unique materials.

\subsubsection*{Data and code availability}
\noindent All data needed to evaluate the conclusions are present in the paper.
Additional data related to this paper may be requested from the lead contact.
The code used for this study is available at \href{https://github.com/zhaoyj21/FCG}{https://github.com/zhaoyj21/FCG}.

\subsection*{Modeling fatigue crack growth}
\noindent To construct the digital libraries of fatigue crack patterns, we consider a linear elastic material for the sake of simplicity.
Young's modulus $Y=200$ GPa and Poisson's ratio $\nu=0.31$ (e.g., for a typical nickel alloy) are chosen as a representative example~\cite{basrour2000Ni}.
XFEM is used to model FCG under uniaxial tension and additional shear components.
In XFEM, special functions are added to the continuous displacement fields to model discontinuity problems such as cracks and interface, which solves difficulties in meshing high-stress areas near the crack tip in FEA~\cite{moes1999XFEM}.
The step of crack advancement is set to $0.3\ {\rm mm}$, which is minute enough in comparison to the size of the structural components ($10\ {\rm mm}$) to ensure convergence.
The angle of crack deflection ($\theta$) is determined by the maximum tangential stress criterion~\cite{zhang2016crack},
\begin{equation}\label{eq1}
	\theta = \arccos{\frac{3K_{\rm \uppercase\expandafter{\romannumeral2}}^2+\sqrt{K_{\rm \uppercase\expandafter{\romannumeral1}}^4+8K_{\rm \uppercase\expandafter{\romannumeral1}}^2K_{\rm \uppercase\expandafter{\romannumeral2}}^2}}{K_{\rm \uppercase\expandafter{\romannumeral1}}^2+9K_{\rm \uppercase\expandafter{\romannumeral2}}^2}},
\end{equation}
where $K_{\rm \uppercase\expandafter{\romannumeral1}}$ and $K_{\rm \uppercase\expandafter{\romannumeral2}}$ are mode-$\rm \uppercase\expandafter{\romannumeral1}$ and the $\rm \uppercase\expandafter{\romannumeral2}$ stress intensity factors (SIFs), respectively.
The remaining life of structures with an existing crack is evaluated by integration using the Paris-Erdogan equation~\cite{paris1963Paris},
\begin{equation}\label{eq1}
	{\rm d}a/{\rm d}N = C(\Delta K)^m,
\end{equation}
where $\Delta K$ is the difference between the maximum and minimum SIFs in a load cycle, $C = 9.7\times10^{-12}$ and $m = 3.0$ are material coefficients~\cite{park2022C_m_para}.

A path-slicing technique is proposed for the uncertainties in service conditions, where the structure is discretized into $N_{\rm s}$ segments.
The loading conditions change across their boundaries, which follow the distribution of statistical noises (assumed to be Gaussian, \zfig{Figure 3A}).
Tails of the Gaussian distribution with a relative probability below $0.05$ are considered as rare events.
The produced digital libraries contain $1,000$ voxel datasets of the crack patterns and their corresponding life to failure.

\subsection*{Model reduction}
\noindent Fatigue cracks can be represented as curves or surfaces for 2D or 3D models.
In this work, we use VAE for nonlinear dimensionality reduction into the latent representations, where the fatigue crack patterns in the digital libraries follow Gaussian distributions~\cite{kingma2013VAE}.
The encoder and decoder in VAE approximate the posterior and likelihood distributions, respectively,
\begin{equation}\label{eq1}
	p(\mathbf{z}\mid \mathbf{x}) = \frac{p(\mathbf{x}\mid \mathbf{z})p(\mathbf{z})}{p(\mathbf{x})},
\end{equation}
where $p(\mathbf{z}\mid \mathbf{x})$ is the posterior, $p(\mathbf{x}\mid \mathbf{z})$ is the likelihood, and $p(\mathbf{z})$ is the prior distribution, $\mathbf{x}$ is the fatigue crack and $\mathbf{z}$ is the latent representation in reduced dimension.
The high-dimensional data is fed into the encoder to reduce the dimensionality.
The decoder then recovers the high-dimensional data from the latent reduced representation.
The loss function of the training process is defined by the reconstruction errors and the KL divergence~\cite{kingma2013VAE},
\begin{equation}\label{eq1}
	{\rm Loss} = \Vert \mathbf{x} - \mathbf{\widehat{x}} \Vert + {\rm KL}(\mathcal{N}(\mathbf{\mu},\bf{\Sigma})\Vert \mathcal{N}(\bf{0},\bf{I})),
\end{equation}
\begin{equation}\label{eq1}
	{\rm KL}(\mathcal{N}(\mathbf{\mu},\bf{\Sigma})\Vert \mathcal{N}(\bf{0},\bf{I})) = \int \mathcal{N}(\mathbf{\mu},\bf{\Sigma})\ln \frac{\mathcal{N}(\mathbf{\mu},\bf{\Sigma})}{\mathcal{N}(\bf{0},\bf{I})}{\rm d}\mathbf{z},
\end{equation}
where $\mathbf{\widehat{x}}$ is the reconstruction of the fatigue crack, $\mathbf{\mu}$ and $\mathbf{\Sigma}$ are the mean and covariance matrix of latent representations learned by neural networks. 
The latent representations in reduced dimension are used to learn the history or path dependence and the nonlinear mapping in downstream tasks.

\subsection*{Statistical learning}
\noindent FCG depends on the loading conditions and the instantaneous crack configuration with loading-history or crack-path dependence.
LSTM and FNN are integrated to achieve efficient prediction of the path and life of FCG at the structural level.
LSTM is used to train and predict FCG from the observed fatigue crack patterns.
The encoder and decoder both have two LSTM layers followed by a fully connected layer.
There are $100$ neural units in each layer unless otherwise noted.
The Adam optimizer is adopted to update the model parameters in training with hyperparameters, $\eta = 10^{-3}$ (the learning rate), $\beta_1 = 0.9$, $\beta_1 = 0.99$, and $\epsilon = 10^{-7}$~\cite{kingma2014adam}.
A re-weighting technique is proposed here for the rare events, where the weight of fatigue cracks with low probability is boosted by $500$ times.
The predicted growing crack patterns are fed to FNN to forecast the remaining life.

\subsection*{Verification and validation}
\noindent The root mean square error (RMSE) between the predicted crack paths ($\mathbf{r}^{\rm pred}$) and the ground truth ($\mathbf{r}^{\rm truth}$) obtained from FEA is defined as
\begin{equation}\label{eq1}
	{\rm RMSE} = \sqrt{\sum_{i=k}^{n}\frac{(\mathbf{r}^{\rm pred}_{i}-\mathbf{r}^{\rm truth}_{i})^2}{n-k+1}},
\end{equation}
where the full path of a crack is discretized into $n$ points and points $1-k$ are known from observation.
The structural similarity (SSIM) is calculated as
\begin{equation}\label{eq2}
	{\rm SSIM} = \frac{(2\mu_{\rm p}\mu_{\rm t}+c_1)(2\sigma_{\rm pt}+c_2)}{(\mu_{\rm p}^2+\mu_{\rm t}^2+c_1)(\sigma_{\rm p}^2+\sigma_{\rm t}^2+c_2)},
\end{equation}
where $\mu_{\rm p}$ and $\mu_{\rm t}$ are the mean values of the prediction and ground-truth voxel data, respectively.
$\sigma_{\rm p}$ and $\sigma_{\rm t}$ are their variances, and $\sigma_{\rm pt}$ is their covariance.
$c_1 = (0.01R)^2$ and $c_2 = (0.03R)^2$ are parameters defined by the range of the voxel values, $R$~\cite{wang2004SSIM,yang2021PATTERNS}.
A low RMSE value or a high SSIM score indicates a high accuracy of prediction.

\subsection*{The complexity of loading profiles}

\noindent The complexity of loading profiles with various levels of statistical noises and rare events is measured by employing the symbolic
aggregate approximation (SAX)~\cite{lin2003SAX}.
SAX transforms time series data such as the loading profiles into symbolic representations (e.g., words) using piecewise aggregate approximation (PAA) and discretization~\cite{lin2003SAX}.
The loading profile within a time span of $m$ ($t_1, t_2, \cdots, t_m$) can be represented by a vector $\mathbf{A}$ in a reduced $w$-dimensional space according to PAA, that is
\begin{equation}\label{eq4}
	a_i = \frac{w}{m}\sum_{j=\frac{m}{w}(i-1)+1}^{\frac{m}{w}i}t_j,
\end{equation}
\begin{equation}\label{eq3}
	\mathbf{A} = [a_1, a_2, \cdots, a_w].
\end{equation}
$a_i$ ($i = 1,2, \cdots, w$) are then discretized to letters ($\widehat{a}_i$) into $l$ ($ = 10$) equally sized regions according to the value of $a_{i}$, that is
\begin{equation}\label{eq6}
	\widehat{a}_i = {\rm letter}_j,\ {\rm if}\ \beta_{j-1} \leq {a}_i < \beta_{j},
\end{equation}
where $\beta_{j-1}$ ($j = 1,2, \cdots, l$).
The loading profile is then represented as a word $\widehat{A}$ with $w$ letters.
\begin{equation}\label{eq5}
	\widehat{A} = \widehat{a}_1, \widehat{a}_2, \cdots, \widehat{a}_w,
\end{equation}
and data complexity of the loading profile can be estimated by the number of possible words as
\begin{equation}\label{eq7}
	{\rm data\ complexity} = l^w,
\end{equation}
and $l$ is the size of the alphabet.
To ensure the accuracy of PAA, we choose large values of $w$ as the complexity of the time series increases, which are $1,\ 4,\ 8, \ 9$ for constant loads, weak statistical noises, strong statistical noises, and rare events, respectively.

\section*{Acknowledgments}

\noindent This study was supported by the National Natural Science Foundation of China through grants 11825203, 11832010, 11921002, 52090032, 12122204, and 11872150.
The computation was performed on the Explorer 100 cluster system of the Tsinghua National Laboratory for Information Science and Technology.

\section*{Author contributions}

\noindent Z.X. conceived and supervised the research.
Y.Z. and Y.L. performed the finite element simulations and analysis.
Y.Z. developed the statistical learning codes.
All authors wrote the manuscript.

\section*{Declaration of interests}

\noindent The authors declare that they have no competing financial interests. 

\clearpage
\newpage

\bibliographystyle{nexus}
\bibliography{main_text}

\clearpage
\newpage

\section*{Figure legends}

\vspace*{0.2in}

\noindent \textbf{Figure 1. The landscape of fatigue life prediction (FLP) in engineering}

\noindent Empirical models can be constructed by fitting to experimental data.
Finite element analysis (FEA) can solve fatigue crack growth (FCG) problems with complex geometries.
Statistical learning models equipped with path slicing and re-weighting can handle statistical noises and rare events based on physical simulation data.
Digital twins allow real-time structural health monitoring (SHM), FLP, and maintenance management decision-making.

\vspace*{0.2in}

\noindent \textbf{Figure 2. The statistical learning framework for FLP}

\noindent The framework contains three modules.
FEA is utilized to model the complexity of FCG by considering the effect of statistical noises and rare events in the construction of digital libraries.
FEA data are fed to dimensionality reduction and neural network architectures in model training.
FLP is achieved by utilizing the well-trained model in engineering applications.

\vspace*{0.2in}

\noindent \textbf{Figure 3. Statistical noises and rare events in the loading conditions}

\noindent (\textbf{A})  The procedures of path slicing and re-weighting used to handle the uncertainties in loading profiles.
\noindent (\textbf{B}) The distribution of fatigue crack patterns in the space of major features extracted by variational autoencoder (VAE).
\noindent (\textbf{C}) Unsupervised clustering of fatigue crack patterns based on the probability density of fatigue crack patterns.

\vspace*{0.2in}

\noindent \textbf{Figure 4. Performance of path slicing and re-weighting}

\noindent (\textbf{A}) Fatigue crack path predictions for a flat plate without path slicing and re-weighting.
The predicted results are not corrected along with the observation time, $t_{\rm obs}$.
\noindent (\textbf{B}) Predictions with path slicing, which is corrected by observation data as $t_{\rm obs}$ increases (shown by gradual shades).
\noindent (\textbf{C}) Predictions with path slicing and re-weighting, which are corrected by observation data and offer accurate forecasting by considering the effects of statistical noises and rare events.
\noindent (\textbf{D}) The comparison of prediction errors by different models.

\vspace*{0.2in}

\noindent \textbf{Figure 5. Performance assessment of FLP}

\noindent (\textbf{A}) Schematic diagram for calculating the structural similarity (SSIM) of fatigue cracks in a flat plate.
\noindent (\textbf{B}) The average accuracy of FLP measured by SSIM for all samples in the test set, which increases with $t_{\rm obs}$.
\noindent (\textbf{C}) The predicted life, the ground truth, and the accuracy of a representative sample in the test set.
The accuracy is defined as $1-\vert \tau_i-\widehat{\tau}_i \vert/\sum_{i} \vert \tau_i-\widehat{\tau}_i \vert$, where $\tau_i$ and $\widehat{\tau}_i$ are the ground truth and predictions of fatigue life with $t_{\rm obs}$.
\noindent (\textbf{D}) The average error in FLP measured by root mean square error (RMSE) for all samples in the test set.
The RMSE decreases with $t_{\rm obs}$.
The standard deviations are reported in the error bars.

\vspace*{0.2in}

\noindent \textbf{Figure 6. Reduction of the loading-profile complexity}

\noindent (\textbf{A}) FLP with path-slicing and re-weighting under loading conditions with continuous profiles for a flat plate.
\noindent (\textbf{B}) The convergence analysis of the size of training sets and $N_{\rm s}$, which is measured by SSIM.
\noindent (\textbf{C}) The complexity of loading profiles with various strengths of uncertainties, which is significantly reduced by introducing path slicing and re-weighting.
\noindent (\textbf{D}) FCG in turbine blades with a complex 3D geometry.

\clearpage
\newpage

\begin{figure}[htp]
\centering
\includegraphics[width=\linewidth] {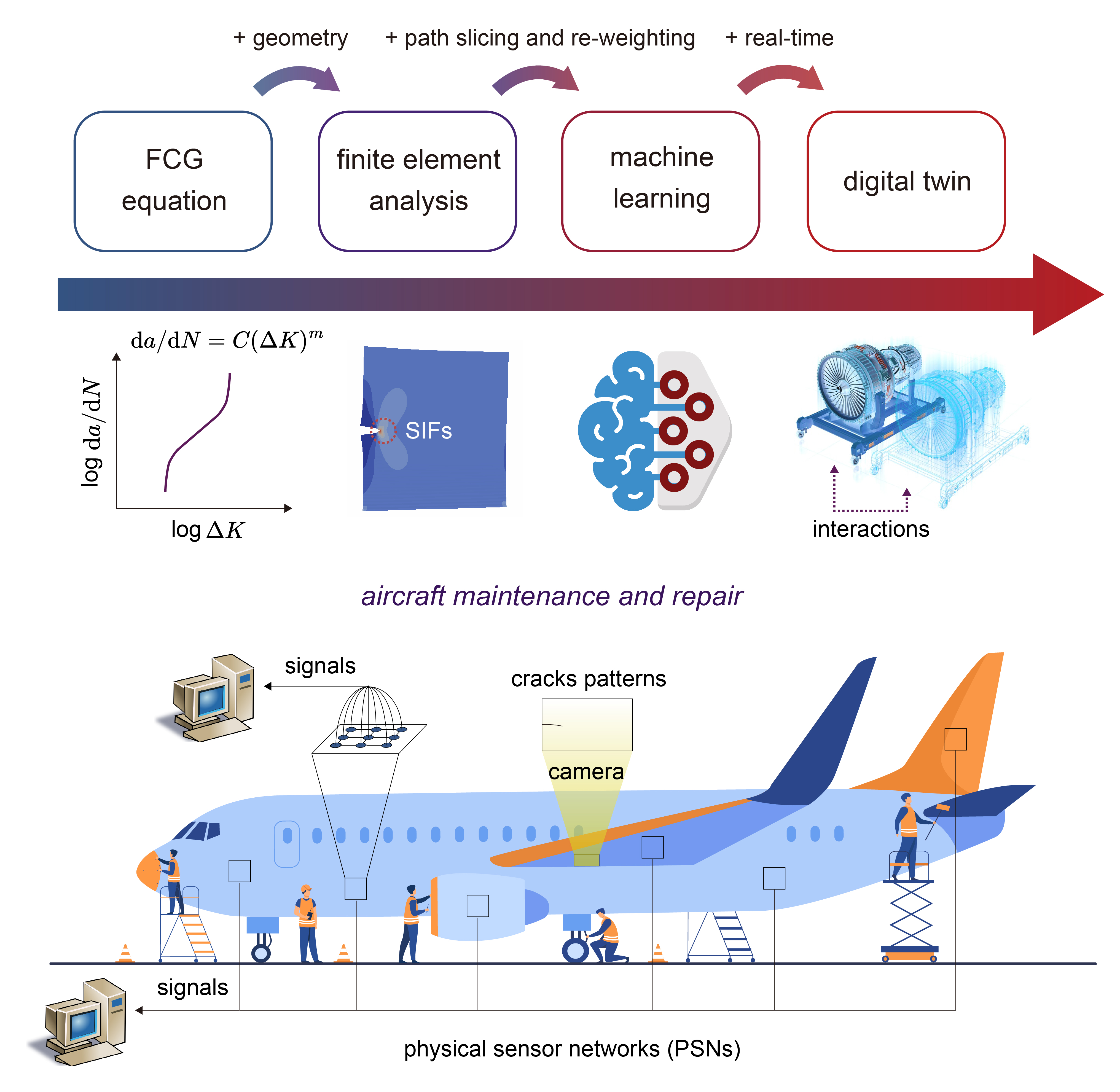}
\label {fig1}
\end{figure}

\clearpage
\newpage
\begin{figure}[htp]
\centering
\includegraphics[width=\linewidth] {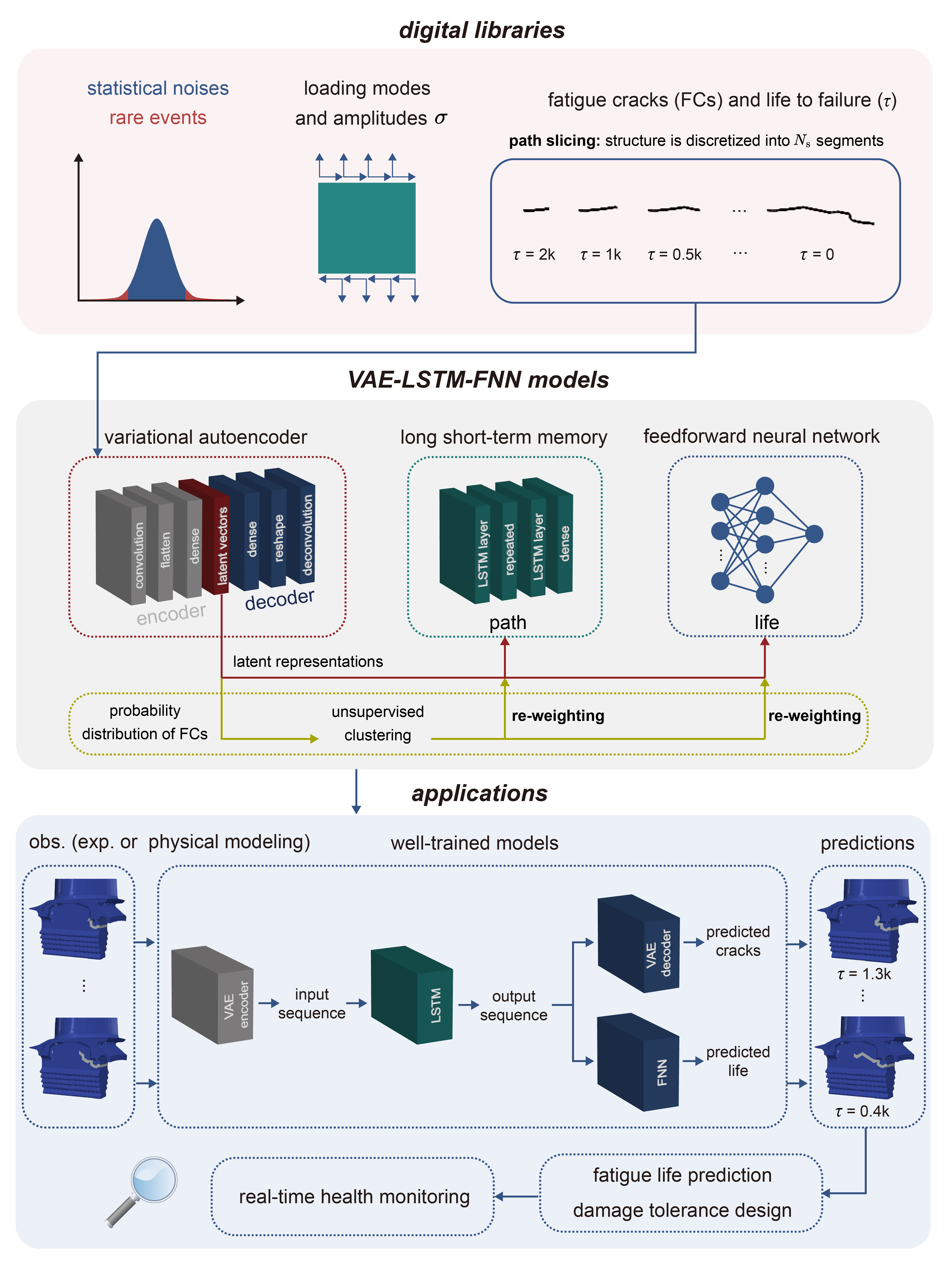}
\label {fig2}
\end{figure}

\clearpage
\newpage
\begin{figure}[htp]
\centering
\includegraphics[width=\linewidth] {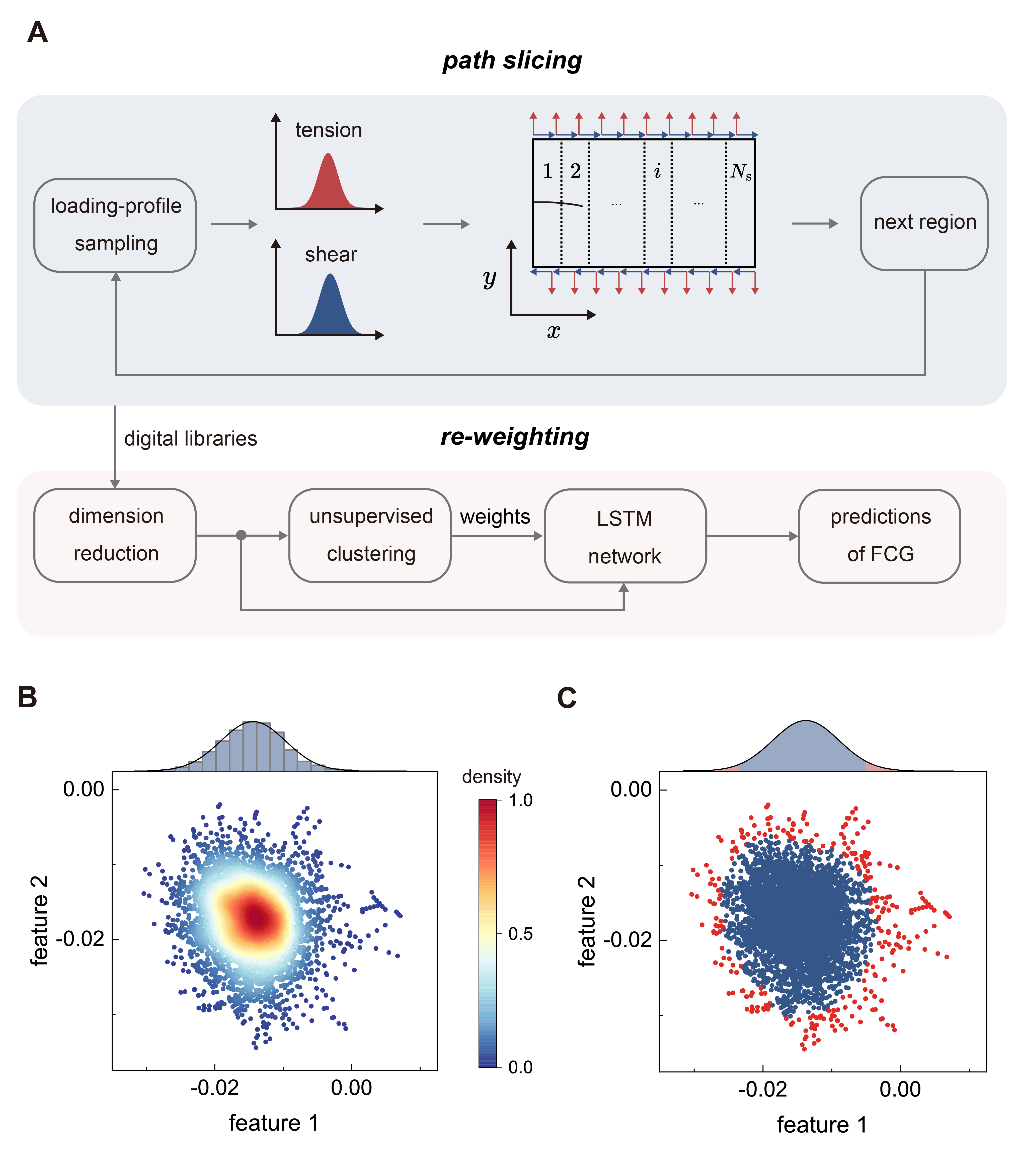} 
\label {fig3}
\end{figure}

\clearpage
\newpage
\begin{figure}[htb]
\centering
\includegraphics[width=\linewidth] {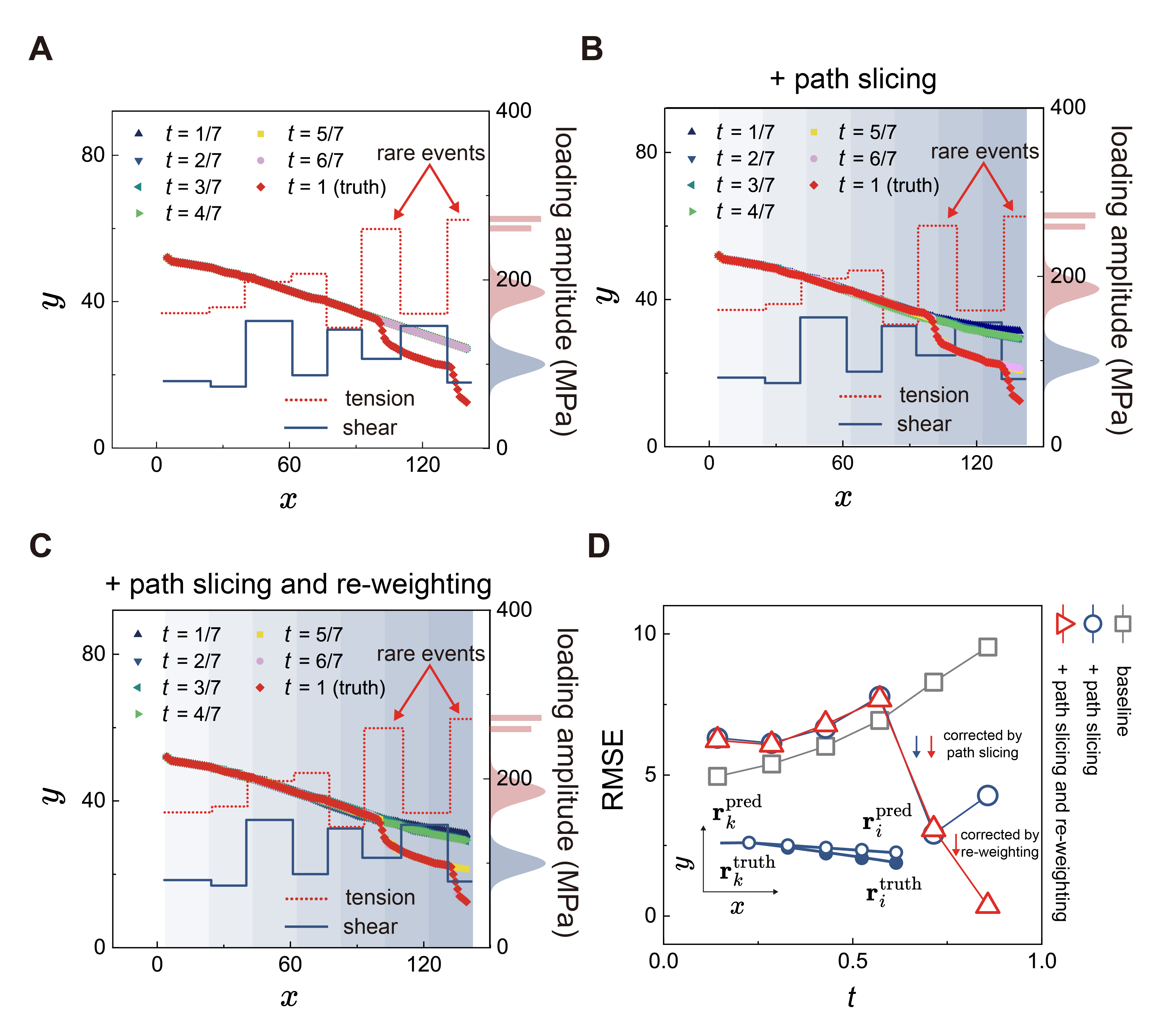} 
\label {fig4}
\end{figure}

\clearpage
\newpage
\begin{figure}[htb]
\centering
\includegraphics[width=\linewidth] {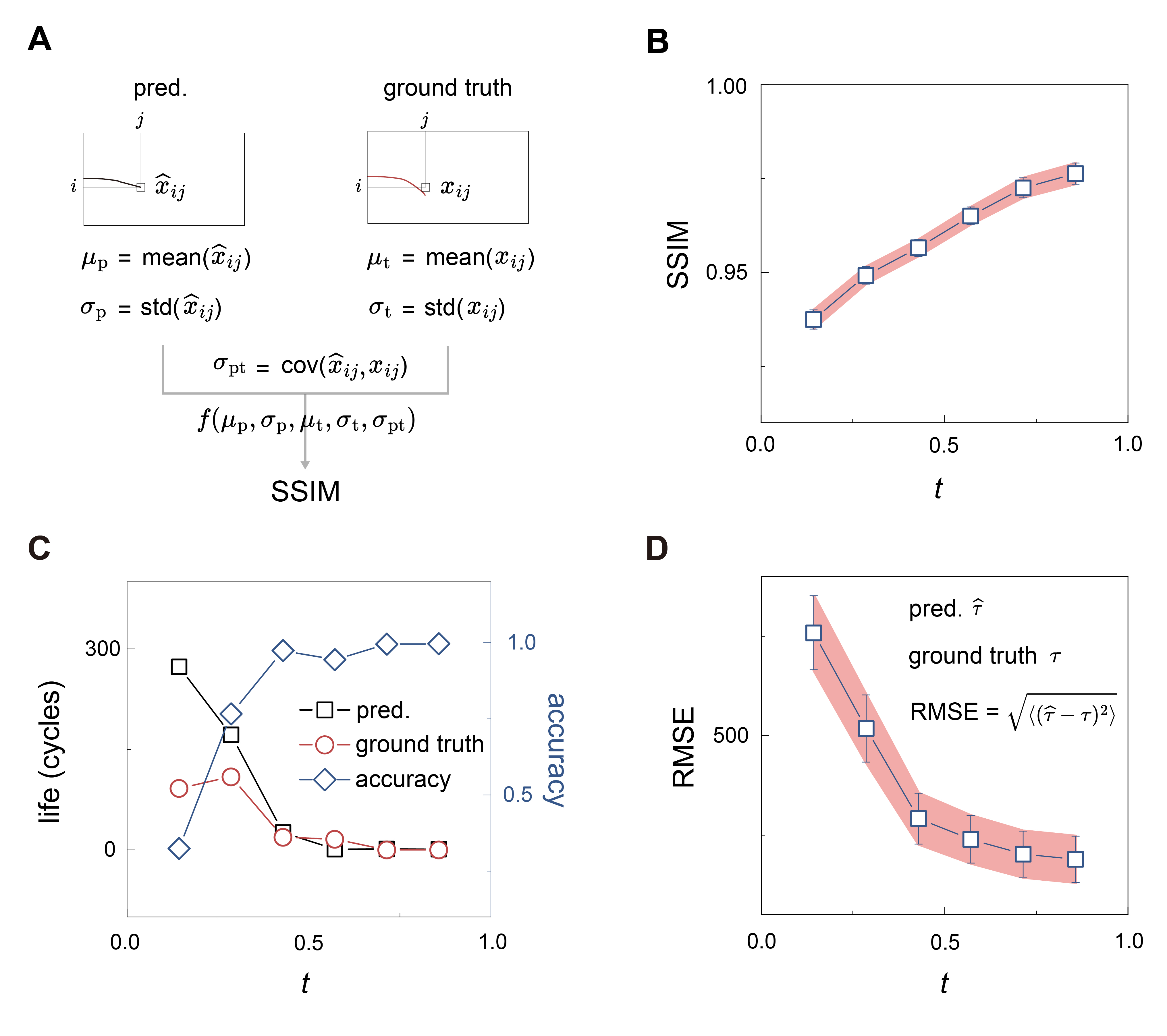} 
\label {fig5}
\end{figure}

\clearpage
\newpage
\begin{figure}[htb]
\centering
\includegraphics[width=\linewidth] {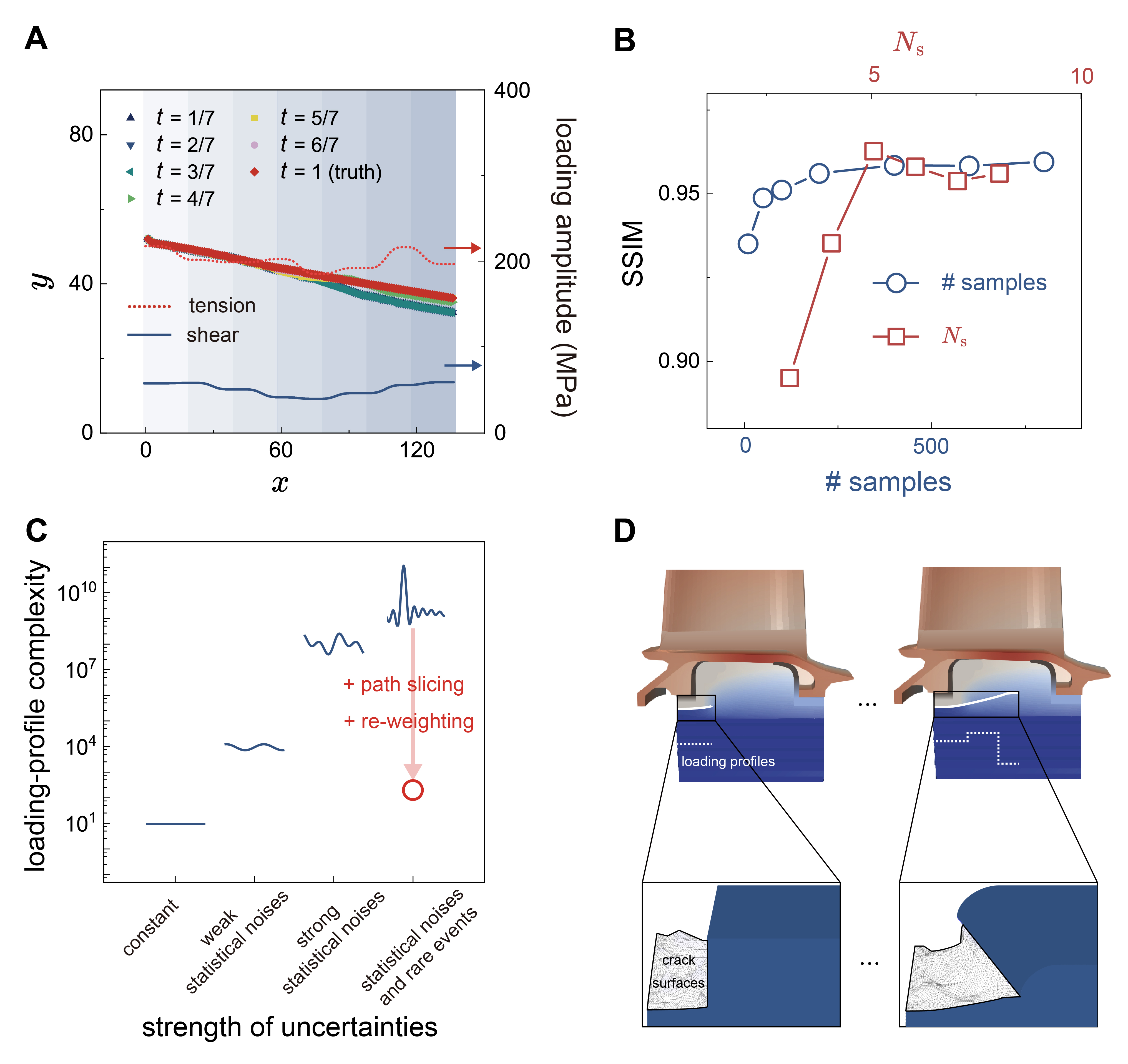} 
\label {fig6}
\end{figure}

\end{document}